\documentclass{article}
\usepackage{ijcai22}

\usepackage{times}
\usepackage{soul}
\usepackage{url}
\usepackage[hidelinks]{hyperref}
\usepackage[utf8]{inputenc}
\usepackage[small]{caption}
\usepackage{graphicx}
\usepackage{amsmath}
\usepackage{amssymb}
\usepackage{amsthm}
\usepackage{booktabs}
\usepackage{algorithm}
\usepackage{algorithmic}
\usepackage{multirow}
\usepackage[usenames,dvipsnames]{color}
\usepackage{colortbl}
\usepackage[dvipsnames]{xcolor}
\usepackage{bbm}
\usepackage[misc]{ifsym} 
\title{Region-Aware Metric Learning for Open World Semantic Segmentation via Meta-Channel Aggregation}

\author{
Hexin Dong$^{1*}$\and Zifan Chen$^{1*}$\and Mingze Yuan$^1$\and Yutong Xie$^1$\and Jie Zhao$^{1,2}$\and Fei Yu$^1$ \and \\ 
 Bin Dong$^{4,3,2}$\And Li Zhang$^{1,2}$$^{(\textrm{\Letter})}$
\affiliations
$^1$Center for Data Science, Peking University, Beijing, China\\
$^2$National Biomedical Imaging Center, Peking University, Beijing, China\\
$^3$Center for Machine Learning Research, Peking University, Beijing, China\\
$^4$Beijing International Center for Mathematical Research (BICMR), Peking University, Beijing, China\\
}

\begin{document}

\maketitle
\begin{abstract}
As one of the most challenging and practical segmentation tasks, open-world semantic segmentation requires the model to segment the anomaly regions in the images and incrementally learn to segment out-of-distribution (OOD) objects, especially under a few-shot condition. The current state-of-the-art (SOTA) method, Deep Metric Learning Network (DMLNet), relies on pixel-level metric learning, with which the identification of similar regions having different semantics is difficult. Therefore, we propose a method called region-aware metric learning (RAML), which first separates the regions of the images and generates region-aware features for further metric learning. RAML improves the integrity of the segmented anomaly regions. Moreover, we propose a novel meta-channel aggregation (MCA) module to further separate anomaly regions, forming high-quality sub-region candidates and thereby improving the model performance for OOD objects. To evaluate the proposed RAML, we have conducted extensive experiments and ablation studies on \textit{Lost And Found} and \textit{Road Anomaly} datasets for anomaly segmentation and the \textit{CityScapes} dataset for incremental few-shot learning. The results show that the proposed RAML achieves SOTA performance in both stages of open world segmentation. Our code and appendix are available at \href{https://github.com/czifan/RAML}{https://github.com/czifan/RAML}.


\end{abstract}
\footnote{$^*$ Equal contribution.}
\footnote{$^{\textrm{\Letter}}$Correspondence to Li Zhang:  zhangli\_pku@pku.edu.cn}
\section{Introduction}

\begin{figure}
	\centering
	\includegraphics[width=0.48\textwidth]{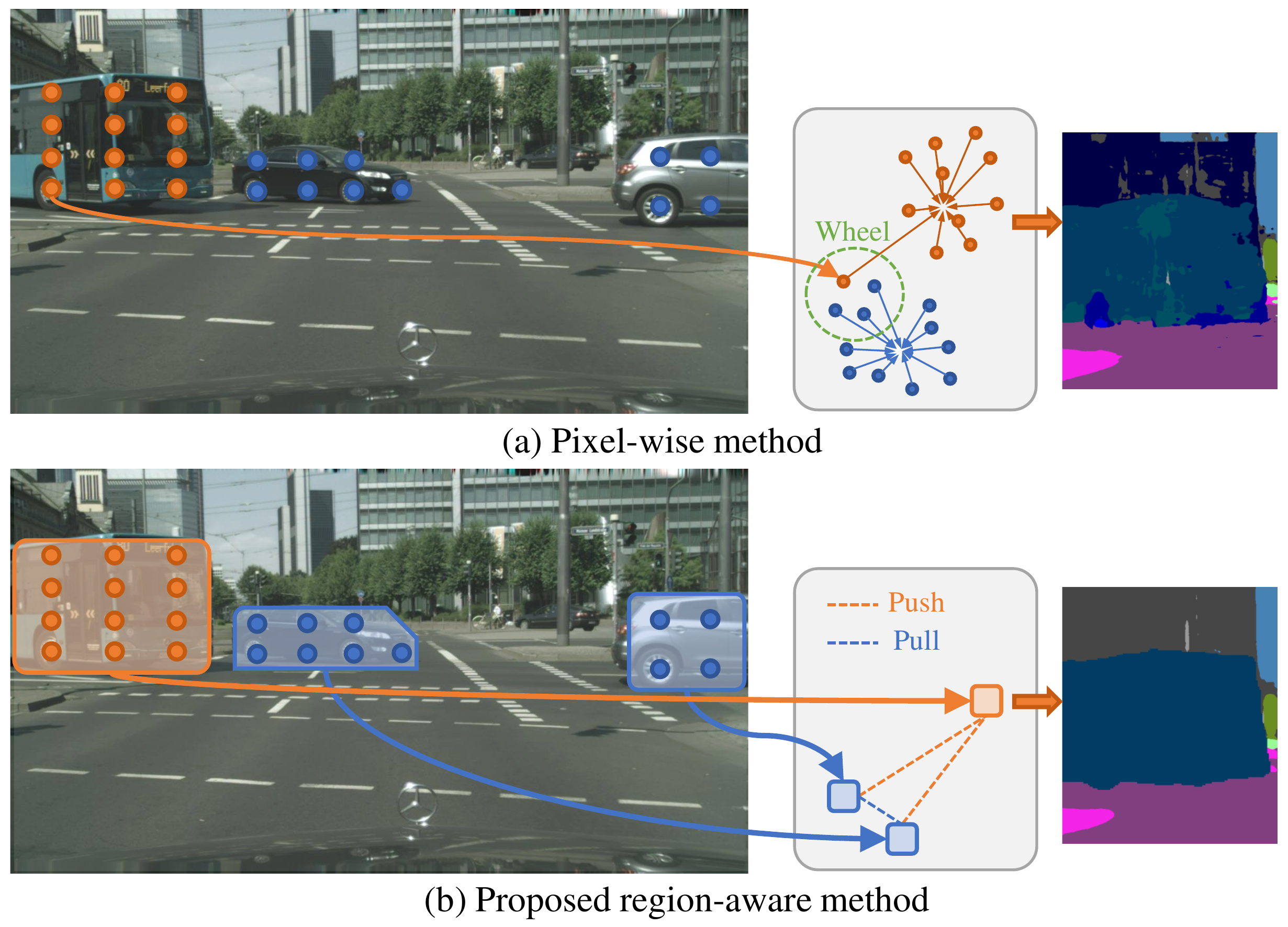}
	\caption{Main idea of our proposed method. (a) Existing methods focus on pixel-level which may result in fine-grained segmentation errors. (b) Our proposed Region-Aware Metric Learning (RAML) method maintains the semantic integrity of the OOD objects.}
	\label{fig:main idea}
\end{figure}

The breakthrough of deep learning in many fields of computer vision is based on the closed set assumption, which means that all classes in the test should be covered in the training set. However, this assumption rarely holds in the open world. Since most computer vision applications have to deal with unknown classes, models, especially the deep models, must handle the out-of-distribution (OOD) data. Quite a number of work for image recognition and classification in the open world has been proposed since the first introduction of the concept ``open world'' in \cite{7298799}. However, the work about open world segmentation is scarce. It is not until recently that \cite{cen2021deep} proposes a two-step framework to achieve open world semantic segmentation. The framework consists of (1) an \textbf{anomaly segmentation} module that extends the close-set model of in-distribution objects to delineate the unknown regions of the OOD objects correctly, and (2) an \textbf{incremental few-shot learning} module that separates the unknown regions into OOD objects with novel classes. They also introduce metric learning into both stages of open world segmentation, and the results prove that their proposed criteria of metric learning can improve the model's segmentation of OOD objects. 

Although this pilot work provides a good framework for open world segmentation tasks, the model can be improved in two aspects for better performance. First, the metric learning in \cite{cen2021deep} relies on the pixel-wise feature embeddings, which may falsely split the object into pieces and result in numerous fine-grained segmentation errors. For example, as shown in Figure \ref{fig:main idea}, the \textit{bus wheels} and the \textit{car wheels} have similar feature embeddings and are highly likely to be classified into one group according to the pixel-wise feature embeddings, but they apparently belong to different classes in semantic segmentation. To solve this kind of problems, we propose region-aware metric learning (RAML) for open world segmentation, which significantly outperforms pixel-wise metric learning (PML) in multiple experiments. 

Moreover, we improve the model performance, especially for the incremental few-shot learning stage, by introducing a novel region separation module named meta-channel aggregation (MCA). MCA first aims at over-segmenting the unknown regions into several meta channels. Regions belonging to different meta channels are aggregated to form a segmentation of the objects and then evaluated by the Region-aware Metric Learning module. 

In addition, \cite{cen2021deep} sets a fixed center embedding for each in-distribution class, i.e., a one-hot vector in the feature space. Although the fixed center embedding can effectively create a distance between the distribution of different classes, it ignores the relative similarity between them. For example, in the \textit{Cityscapes} dataset, the method fails to reveal that the difference between \textit{person} and \textit{rider} is smaller than the difference between either of them and \textit{sky}. This paper aims to overcome the drawback by exploiting a more natural metric learning to constrain the distance between the inter-class region-aware features. Specifically, we replace the one-hot setting in \cite{cen2021deep} with Circleloss \cite{2020Circle} as the objective of the metric learning, which not only maintains a fine inter-class distance but also shapes the intra-class distribution more concentrated. Experiments show that such division of the feature space is more conducive to segmenting the OOD data.

In summary, we propose a region-aware metric learning method for open world semantic segmentation. Our contributions are as follows:
\begin{itemize}
	\item We propose using the region-aware over pixel-wise features for open world semantic segmentation to ensure better semantic integrity of the segmented OOD objects.
	
	\item We introduce the MCA module as a novel region separation method that suits incremental few-shot learning.
	
	\item We adopt Circleloss \cite{2020Circle} to enlarge the inter-class distance and reduce the intra-class distance of the data samples, improving the performance of the RAML module.
\end{itemize}

\section{Related Work}
\subsection{Region-aware Semantic Segmentation}
The ideas of how to apply regional information to improve semantic segmentation have been discussed by many research groups recently, including two main threads. First, several works have shown that region-aware information has better contextual representation than pixel-level information to achieve pixel labeling~\cite{yuan2021segmentation}. Secondly, for image segmentation tasks, region-aware information can be better combined with metric or contrastive learning to manipulate the feature space more effectively~\cite{wang2021exploring,hu2021region}. These ideas inspire our paper, but the above works require a sufficient number of training samples to obtain the reasonable region-aware feature representation, while our work is in an open world setting that can only access a few images with unseen class labels. Therefore, we have to design novel region-separation modules (such as MCA) that fit the open world segmentation tasks.
\subsection{Anomaly Segmentation}
There are two types of approaches for anomaly segmentation, including uncertainty-based methods and generative model-based methods. Uncertainty refers to the level of not belonging to known classes, widely used to determine abnormal states. The baseline of uncertainty-based methods is maximum softmax probability (MSP) reported by~\cite{hendrycks2016baseline}. \cite{hendrycks2019scaling} then improves MSP using maximum logit (MaxLogit) for better performance on large-scale datasets. Other uncertainty-based methods include using Bayesian neural networks~\cite{gal2016dropout} and maximizing the entropy of OOD objects in the images~\cite{chan2021entropy}. On the other hand, generative model-based methods also perform well, including autoencoder (AE)~\cite{baur2018deep} and GAN-based methods
~\cite{xia2020synthesize}. However, generative models suffer from unstable training and usually have complex network backbones. 

In this work, we follow the idea of MaxLogit and develop our anomaly segmentation based on non-normalized logit.
\subsection{Open World Problem}
\cite{7298799} is the first research that gives the formal definition of ``open world'', i.e., an open world model must incrementally learn and extend its generality, thereby making the objects with novel classes ``known'' to itself. Since then, the research on open world problems has increased, including classification~\cite{Zhong_2021_CVPR}, object detection~\cite{joseph2021open}, instance segmentation~\cite{saito2021learning}, among others. However, it is not until recently that~\cite{cen2021deep} proposes the first framework of open world semantic segmentation. Our work follows the settings in~\cite{cen2021deep} and divides the problem into anomaly segmentation and incremental few-shot learning. However, to ensure semantic integrity and improve the segmentation performance, we use region-aware feature embedding instead of pixel-wise feature extraction in their original method.

\begin{figure*}[htb]
\centering
\includegraphics[width=0.96\textwidth]{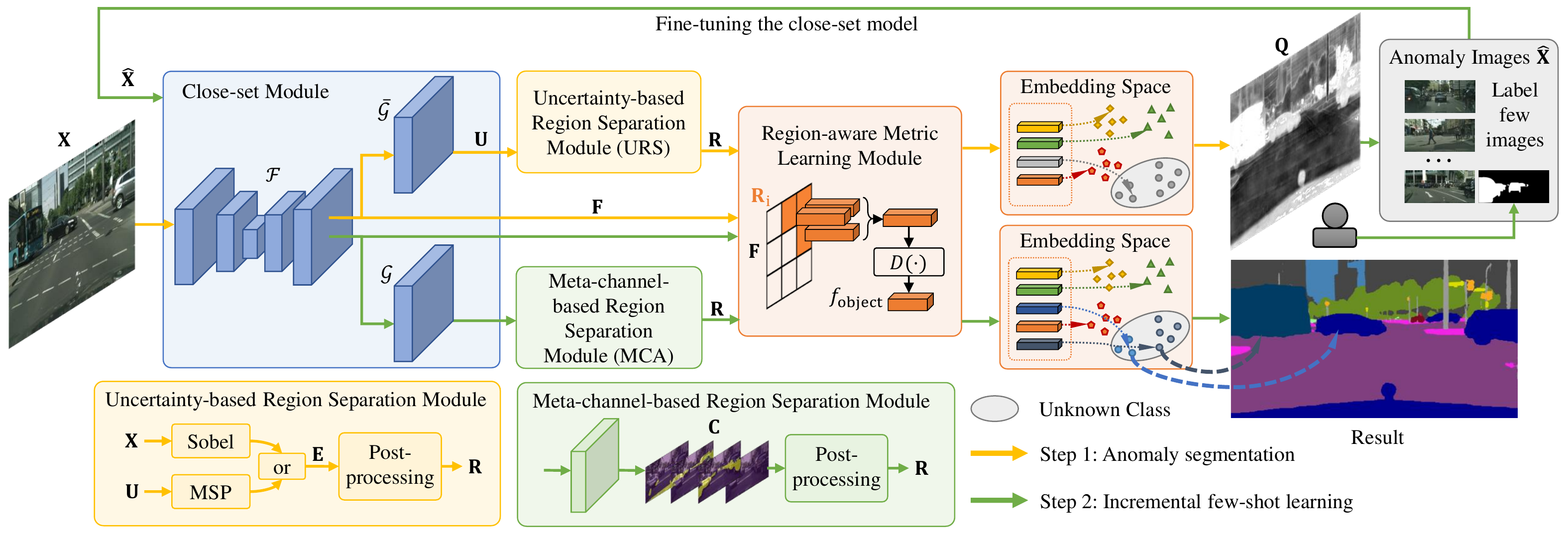}
\caption{The pipeline of Region-aware Metric Learning for Open World Semantic Segmentation: 1) train a close-set segmentation model with known classes (\textit{bluish square}); 2) \textbf{Anomaly Segmentation} (\textit{yellowish arrows}): separate regions based on edge prediction (\textit{yellowish squares}) and segment the anomaly regions using metric learning (\textit{orangish squares}); 3) annotate for unknown objects ($\hat{\mathbf{X}}$) to fine-tune the close-set model; 4) \textbf{Incremental few-shot learning} (\textit{greenish arrows}): separate regions based on MCA (\textit{greenish squares}) and segment the OOD objects using metric learning (\textit{orangish squares}). (Best view in color)}
\label{fig:framework}
\end{figure*}

\subsection{Metric Learning}
Deep metric learning constrains the distance between feature embedding of learning samples to manipulate the feature distribution. Its applications are seen in various computer vision tasks, such as open set recognition~\cite{chen2020learning}, few-shot learning~\cite{oreshkin2018tadam} and open world semantic segmentation~\cite{cen2021deep}. Classic metric learning includes two paradigms. The first is to learn with pair-wise labels, under the guidance of triplet loss ~\cite{Schroff_2015_CVPR} and center loss~\cite{wen_center}. The second consists of softmax cross-entropy and variants that train the model with class-level labels. A recently proposed method called Circle loss~\cite{2020Circle} unifies the above two paradigms and forms the feature space with large inter-class distances and small intra-class distances. We thus adopt Circle loss as the key objective of our proposed RAML module.

\section{Methods}
As shown in \autoref{fig:framework}, our proposed method contains: 1) a backbone model for close-set segmentation, 2) an anomaly segmentation process to delineate the unknown regions of OOD data, and 3) an incremental few-shot learning step for splitting the unknown regions into objects with novel classes.
\subsection{Close-set Segmentation Module}
\label{sec:close-set}
Suppose $ \mathcal{C}_{in} =\{C_{in,1}, C_{in,2}, ... C_{in,N}\}$ are $N$ in-distribution
classes, which are all annotated in training datasets, and $ \mathcal{C}_{out} = \{C_{out,1},C_{out,2}, ... C_{out,M}\}$ are $M$ novel classes not involved in the training datasets. Here, the semantic segmentation network $\mathcal{S}$ is divided into a feature extractor $\mathcal{F}$ and a label predictor $\mathcal{G}$, where $\mathcal{S} = \mathcal{G} \circ \mathcal{F}$.

For the close-set segmentation, we minimize the following loss $\mathcal{L}_{seg}(\mathcal{F},\mathcal{G})$ which guides $\mathcal{S}$ to produce a pixel-level segmentation for in-distribution classes.
\begin{equation}
   \mathcal{L}_{seg}(\mathcal{F},\mathcal{G})  = \mathbb{E}_{\mathbf{X}, \mathbf{Y}}(\ell_{ce}(\mathcal{G} \circ \mathcal{F}(\mathbf{X}), \mathbf{Y}))
\label{equ:seg}
\end{equation}
where $\ell_{ce}(\cdot,\cdot)$ indicates the multi-class cross entropy loss, $\mathbf{X} \in \mathbb{R}^{3 \times H \times W}$ is an input image, $\mathbf{Y}$ is the corresponding label. 

After training this module, we obtain a trained feature extractor $\mathcal{F}$ and a trained label predictor $\mathcal{G}$. The feature map $\mathbf{F} = \mathcal{F}(\mathbf{X}) \in \mathbb{R}^{N_1 \times H \times W}$ and the non-normalized logit $\mathbf{U} = \bar{\mathcal{G}}(\mathbf{F}) \in \mathbb{R}^{N \times H \times W}$ can then be generated for in-distribution classes, where $\bar{\mathcal{G}}$ is obtained by removing the softmax layer of $\mathcal{G}$. The feature map $\mathbf{F}$ and the non-normalized logit $\mathbf{U}$ will be used in later modules.

\subsection{Anomaly Segmentation}
\label{sec:open-set}
\begin{figure}
    \centering
    \includegraphics[width=0.42\textwidth]{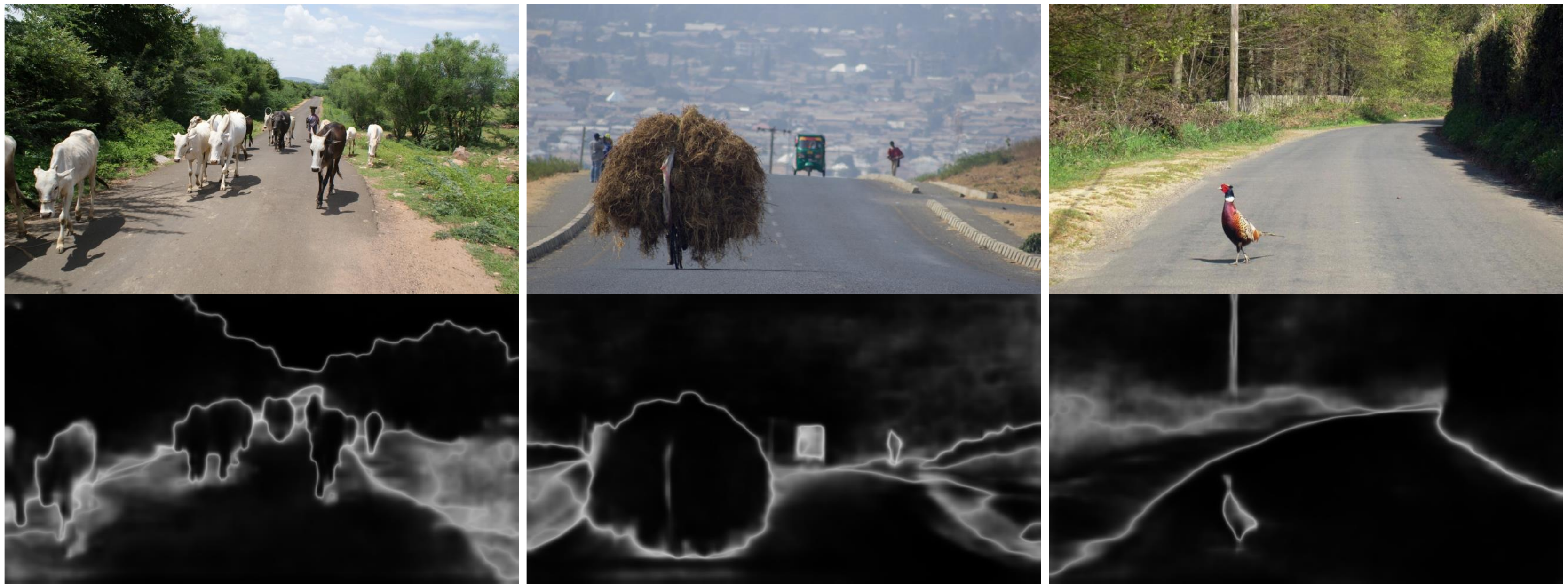}
    \caption{Visual examples of maximum softmax probability. Borders between objects have higher uncertainty because the semantics of the borders are usually ambiguous.}
    \label{fig:msp visulaization}
\end{figure}

To identify the candidate regions of region-aware anomaly segmentation, we adopt an uncertainty-based OOD objects detection method, MSP~\cite{hendrycks2016baseline}, as our region separation module, named Uncertainty-based Region Separation (URS). Its high uncertainty response around the object edges could be used as a promising initialization of the region separation, as shown in Figure~\ref{fig:msp visulaization}.

To further enhance the edges, we introduce Sobel filtering over the original input image. The final edge prediction map $\mathbf{E}$ can be generated as follow,
\begin{equation}
    \label{equ:edge}
    \mathbf{E} = \mathbb{I}\left(\operatorname{Sobel}(\mathbf{X}) \geq \alpha\text{ or }\operatorname{MSP}(\mathbf{U}) \geq \beta \right),
\end{equation}
where $\mathbf{X}$ is the input image, $\mathbf{U}$ is the non-normalized logit, and $\mathbb{I}(\cdot)$ is an indicator function, $\alpha$ and $\beta$ are hyper-parameters to control the edge prediction. According to $\mathbf{E}$, we use a post-processing sub-module, including the hole filling and connected component algorithms, to generate the candidate regions $\mathcal{R} = \{\mathbf{R}_1, \mathbf{R}_2, \cdots, \mathbf{R}_T\}$, where $\mathbf{R}_i\in\{0,1\}^{H \times W}$ represents the $i$-th region. 

We then propose a RAML module for anomaly segmentation to classify the candidate regions $\mathcal{R}$. For each region $\mathbf{R}_i \in \{0, 1\}^{H \times W}$, the region-aware feature embedding is obtained as below:
\begin{equation}
	f_{object} = \mathcal{D}( \frac{ \sum_{j,k}\mathbf{F}^{j,k}\mathbf{R}_i^{j,k} }{\sum_{j,k} \mathbf{R}_i^{j,k}}) \in \mathbb{R}^{N_2}
	\label{equ:object}
\end{equation}
where $\mathbf{F}^{j,k} \in \mathbb{R}^{N_1}$ is the feature vector of pixel $(j, k)$, $\mathcal{D}(\cdot)$ consists of two fully-connected layers to control the embedding dimension.
$f_{object}$ is compared to all the prototypes of the known classes by metric learning constrained by circle loss~\cite{2020Circle}. Specifically, the prototype of $l$-th known class $f_{l}$ can be obtained using the semantic segmentation label. Then, the region-aware anomaly probability of $\mathbf{R}_i$ can be expressed as below,
\begin{equation}
	\mathcal{P} (\mathbf{R}_i,\mathbf{F}) = \max_{1 \leq l \leq N} \frac{f_{object} \cdot {f_l}}{\|f_{object}\| \|f_l\|}.
\end{equation}

Finally, to generate a pixel-level anomalous probability map, we combine the information from the non-normalized logit and the above region-aware anomaly probabilities. For each pixel $(j, k)$, uncertainty intensity $\mathbf{Q}^{j,k}$ is computed as, 
\begin{equation}
    \mathbf{Q}^{j,k}=-\max_{1 \leq l \leq N}\ \mathbf{U}^{j,k}_{(l)} \cdot \mathcal{P}(\mathbf{R}_i, \mathbf{F}),
\end{equation}
where the pixel $(j, k)$ belongs to region $\mathbf{R}_{i}$, $\mathbf{F}$ is the feature map,  $\mathcal{P}(\cdot, \cdot)$ is the region-aware anomaly probabilities. $\mathbf{U}^{j,k}_{(l)}$ is the $l$-th output of pixel $(j,k)$ in the non-normalized logit $\mathbf{U}$. We then normalize the uncertainty intensity $\mathbf{Q}^{j,k}$ for each pixel to obtain the anomalous probability map, which is used to identify the unknown regions in the image. 

\subsection{Incremental Few-shot Learning via MCA}
\label{sec:open-world}
After the anomaly segmentation, open world semantic segmentation requires the model to identify all objects of $M$ novel classes in the unknown regions. One way to realize the incremental few-shot learning is to use a few labeled images containing objects with novel classes to fine-tune the close-set segmentation model under the loss $\mathcal{L}_{seg}$. However, experiments show that this improvement is trivial. We thus propose an innovative MCA module for further creating sub-regions in the unknown regions from anomaly images $\hat{\mathbf{X}}$. 
MCA takes the prediction of the label predictor $\mathcal{G}$ in the close-set model as its input to output $(N+K)$ channels with softmax activation $\mathbf{C} \in [0,1]^{(N+K) \times H \times W}$.
The first $N$ channels are the segmentation results for all in-distribution classes, while the last $K (K>M)$ channels are \textit{meta channels} to overly segment the unknown regions. Several MCA-related losses are integrated into $\mathcal{L}_{seg}$ during the fine-tuning, and the overall loss function is,
\begin{equation}
\mathcal{L}_{overall} = \mathcal{L}_{seg}+ \lambda_{inter} \mathcal{L}_{inter} + \lambda_{split} \mathcal{L}_{split} + \lambda_{rec} \mathcal{L}_{rec}.    
\end{equation}

The first term $\mathcal{L}_{seg}$ is the segmentation loss for all in-distribution classes from \autoref{equ:seg}. The second term utilizes the negative of Dices to minimize the intersection between any pairs of output channels, which is defined as:
\begin{equation}
    \mathcal{L}_{inter} = \sum_{1 \leq i < j \leq N+K} (1-\ell_{dice}(\mathbf{C}_i,\mathbf{C}_j))
\end{equation}
where $\ell_{dice}(\cdot,\cdot)$ indicates the dice loss and $\mathbf{C}_i$, $\mathbf{C}_j$ are the $i$-th and $j$-th channels of the segmentation output. 

The third term aims to avoid the sub-regions (candidates of OOD objects) gathering in a few certain channels:
\begin{equation}
    \mathcal{L}_{split} = \sum_{i=N+1}^{N+K}-\log( \max(\eta \sum_{j,k} \mathbf{C}^{j,k}_{i},1))
\end{equation}
where $\mathbf{C}^{j,k}_{i}$ represent $(j,k)$ pixel output of $i$-th channel and $\eta$ is a hyper-parameter to control the separation. $\mathcal{L}_{split}$ reaches the minimum when the sub-regions scatter across the output channels according to Jenson's inequality.

The last term encourages the outputs of all channels to reconstruct the entire image, further avoiding loss of information: 
\begin{equation}
    \mathcal{L}_{rec} = {||\mathbf{X} \odot ( \sum_{i=1}^{N+K}\mathbf{C}_i - \mathbbm{1}_{H \times W})||^2}
\end{equation}
where $\odot$ is the element-wise multiplication operator and $\mathbbm{1}_{H \times W}$ is a matrix with all ones.
\begin{figure}
    \centering
    \includegraphics[width=0.43\textwidth]{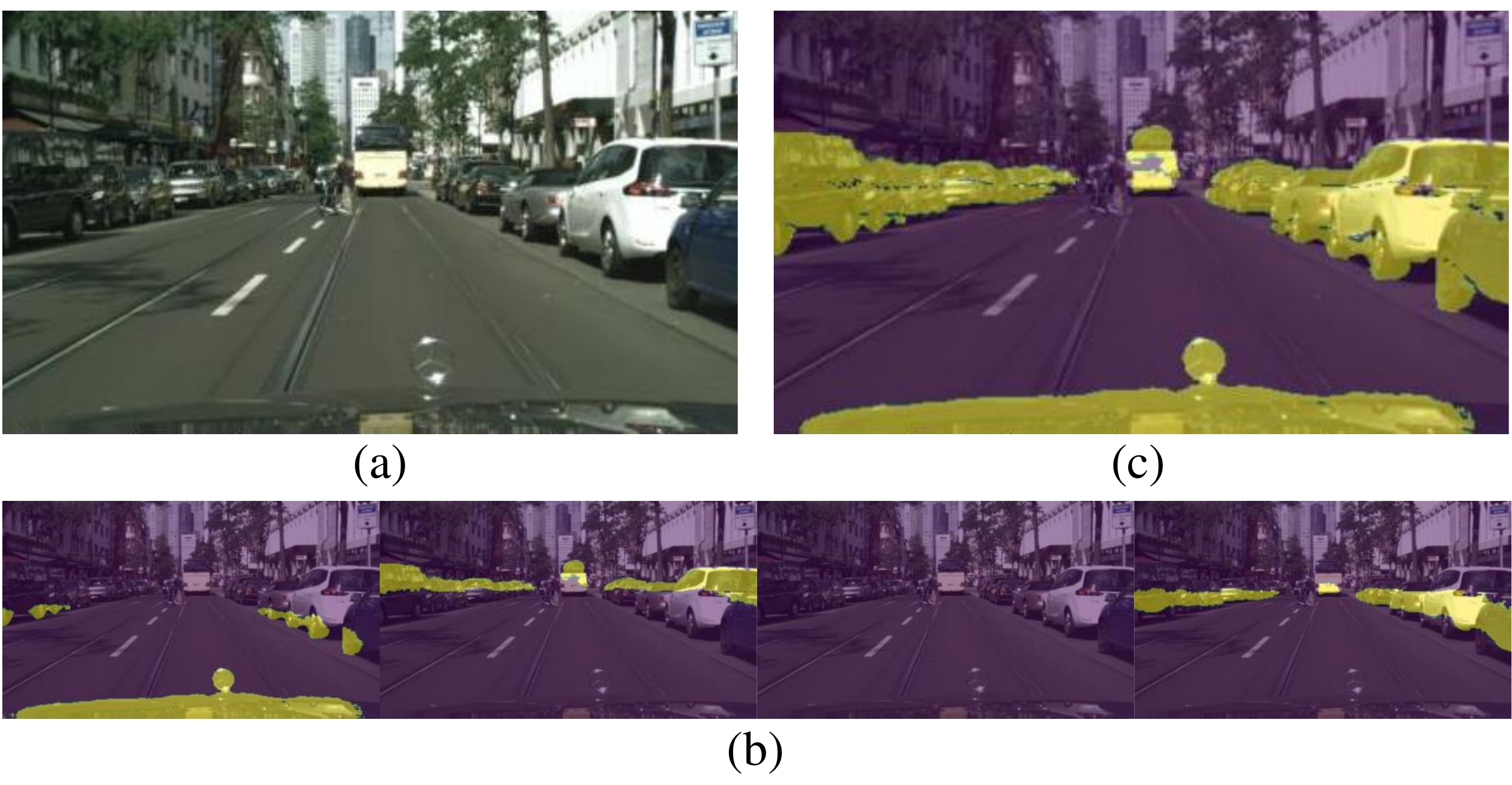}
    \caption{Visualization results of MCA. (a) Input image; (b) Meta-channel response ($K=4$); (c) Aggregated Meta-channel.}
    \label{fig:meta channel}
\end{figure}

As shown in \autoref{fig:meta channel}, we observe that MCA tends to segment objects based on local semantic information. One unknown object may be segmented into more than one channel and lose completeness. (e.g., The windows and wheels of cars may be divided into different channels.) 
Therefore, we aggregate the sub-regions from certain meta channels according to few-shot (here $L$-shot) labeled images, which generates the candidate regions $\mathcal{R} = \{\mathbf{R}_1, \mathbf{R}_2, \cdots, \mathbf{R}_T\}$ for the final RAML module of incremental few-shot learning.

Similar to \autoref{equ:object}, the region-aware feature embedding $f_{object}$ for each region $\mathbf{R}_i$ could be computed. The prototype of $i$-th unknown class ($1 \leq i \leq M$) from $L$-shot newly labeled images is defined as:
\begin{equation}
     c_{i} = \frac{1}{L} \sum_{j=1}^{L} f_{i}^{(j)}
\end{equation}
where $f_{i}^{(j)}$ represents the feature embedding of $i$-th unknown class in $j$-th annotated image. For each region-aware feature embedding $f_{object}$, we use cosine similarity to measure the distance between this candidate region and every unknown class:
\begin{equation}
     s_{object}^{i} = \frac{f_{object} \cdot {c_i}}{\|f_{object}\| \|c_i\|}, i= 1,2,...,M 
\end{equation}
The candidate region can be classified as the $i$-th novel class $C_{out,i}$ only if the cosine similarities satisfy the following two criteria:
\begin{equation}
\label{equ:criteria}
\left\{
     \begin{array}{cc}
          s_{object}^i > \theta_{novel} \\
          s_{object}^i > s_{object}^{i'} &  \forall i'\neq i 
     \end{array}
\right.
\end{equation}
where $\theta_{novel}$ is a hyper-parameter to control classification. 
\section{Experiments}
Our experiments include three parts: (1) experimental results of anomaly segmentation in \autoref{sec:openset_exp}; (2) experimental results of incremental few-shot learning results in \autoref{sec:openworld_exp}; (3) ablation studies in \autoref{sec:ablation} and Appendix.
\subsection{Anomaly Segmentation}
\label{sec:openset_exp}

\begin{figure}
    \centering
    \includegraphics[width=0.48\textwidth]{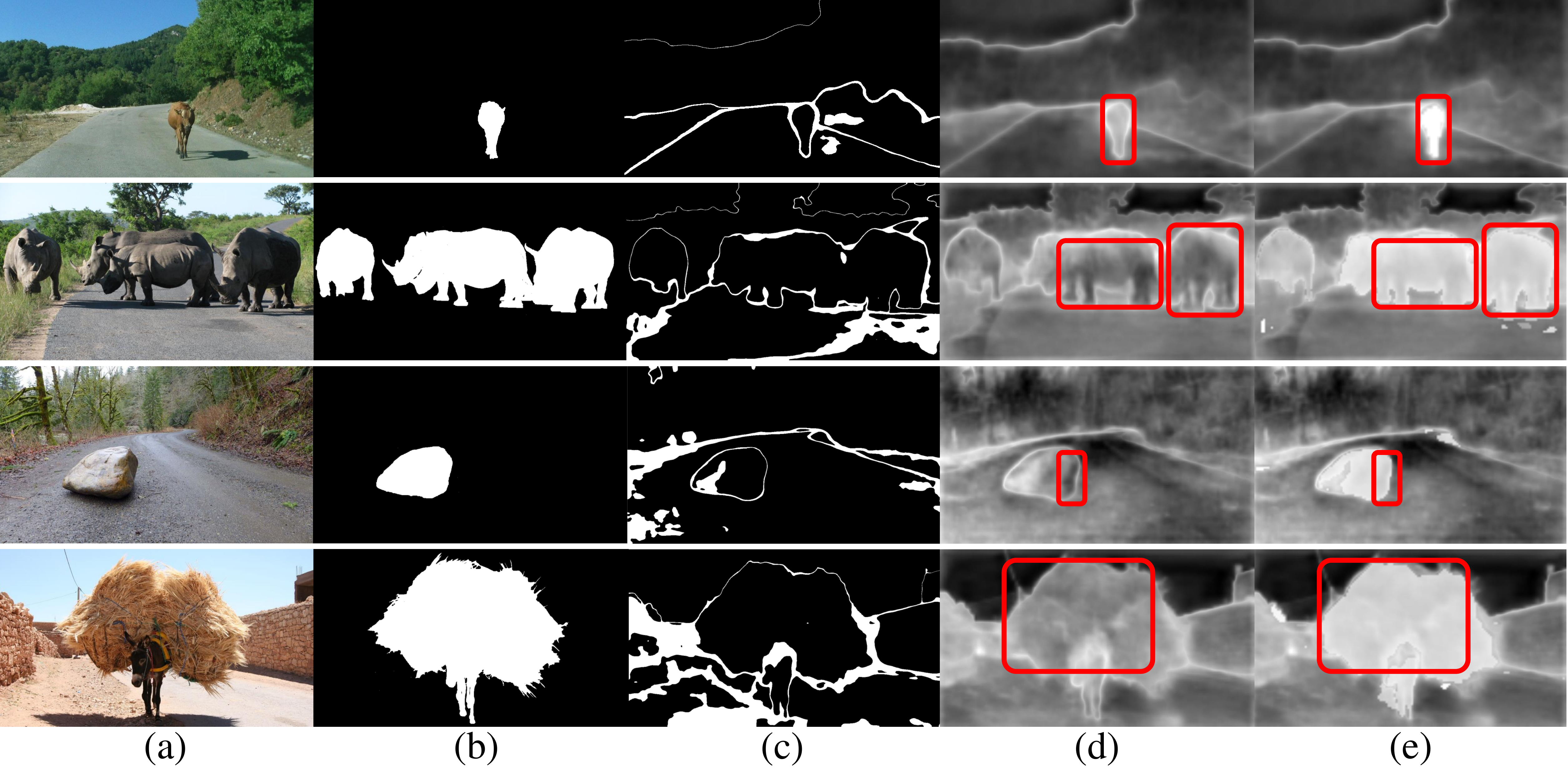}
    \caption{Visualization results of \text{anomaly segmentation} on \textit{Road Anomaly}. (a) input image; (b) ground truth; (c) edge prediction; (d) results of MaxLogit \protect \cite{hendrycks2019scaling}. (e) results of our proposed RAML method. For (d) and (e), higher value represents greater probability of anomaly. The red bounding boxes indicate that RAML ensures the integrity of the anomaly regions.}
    \label{fig:anomaly segmentation}
\end{figure}

 \paragraph{Datasets.} 7000 full-frame annotated driving scenes from \textit{BDD100k}~\cite{yu2020bdd100k} are used to train the close-set segmentation model, containing 19 categories of objects as in-distribution objects. For anomaly segmentation, we use another two road scene datasets, \textit{Lost and Found}~\cite{pinggera2016lost} and \textit{Road Anomaly}~\cite{lis2019detecting}, with anomalous objects other than ones in \textit{BBD100k}.

\paragraph{Implementation details.} We follow~\cite{hendrycks2019scaling,cen2021deep} to use PSPNet as the network backbone of our close-set segmentation module and apply two fully connected layers for RAML. We follow~\cite{hendrycks2016baseline} to use three metrics to evaluate the performance of anomaly segmentation, including area under ROC curve (AUROC),  area under the precision-recall curve (AUPR), and the false-positive rate at 95\% recall (FPR95).


\paragraph{Results.} As shown in Table~\ref{tab:anomaly segmentation}, our proposed RAML module achieves the SOTA performance on \textit{Lost and Found} and \textit{Road Anomaly} for anomaly segmentation. Figure~\ref{fig:anomaly segmentation} presents some visual examples to compare RAML and the pixel-wise method. The proposed RAML module produces higher response values and better integrity within the anomalous objects, significantly reducing the false-negative cases.

\begin{table}
\centering
\resizebox{0.48\textwidth}{!}
{
\begin{tabular}{l|ccc|ccc}
\hline
Dataset  & \multicolumn{3}{|c|}{\textit{Lost and Found}} & \multicolumn{3}{|c}{\textit{Road Anomaly}} \\
\hline
Method & AUPR$\uparrow$ & AUROC$\uparrow$ & FPR95$\downarrow$  & AUPR$\uparrow$ & AUROC$\uparrow$ & FPR95$\downarrow$ \\
\hline
Ensemble & - & 57 & - & - & 67 & - \\
RBM & - & 86 & - & - & 59 & - \\
MSP & 21 & 83 & 31 & 19 & 70 & 61 \\
MaxLogit & 37 & 91 & 21 & 32 & 78 & 49 \\
DUIR & - & 93 & - & - & 83 & - \\
DML & 45 & \textbf{97} & 10 & 37 & 84 & 37 \\
\hline
RAML(Ours) & \textbf{46} & \textbf{97} & \textbf{8} & \textbf{42} & \textbf{86} & \textbf{32} \\
\hline
\end{tabular}
}
\caption{Results of anomaly segmentation on \textit{Lost and Found} and \textit{Road Anomaly}.}
\label{tab:anomaly segmentation}
\end{table}

\subsection{Incremental Few-shot Learning}
\label{sec:openworld_exp}
\paragraph{Datasets.}
we use \textit{Cityscapes} dataset to train and evaluate our RAML module in the incremental few-shot learning step. \textit{Cityscapes} consists of 2975 real-world images in the training set and 500 in the validation set with a resolution of $2048 \times 1024$. The division of training set and test set in our experiments is consistent with this division.

\paragraph{Implementation details.} We follow \cite{cen2021deep} to train a DeeplabV3+ model as the close-set model, which is followed by two fully connected layers for RAML and use mean Intersection-over-Union (mIoU) to evaluate the performance of segmentation results. Specifically, \textbf{mIoU\scriptsize{old}} and \textbf{mIoU\scriptsize{novel}} are the mIoUs of known and unknown classes, respectively. The metric \textbf{mIoU\scriptsize{harm}} is a comprehensive index \cite{2019Semantic} that balances \textbf{mIoU\scriptsize{old}} and \textbf{mIoU\scriptsize{novel}}.


\begin{figure}[h]
\centering
\includegraphics[width=0.48\textwidth]{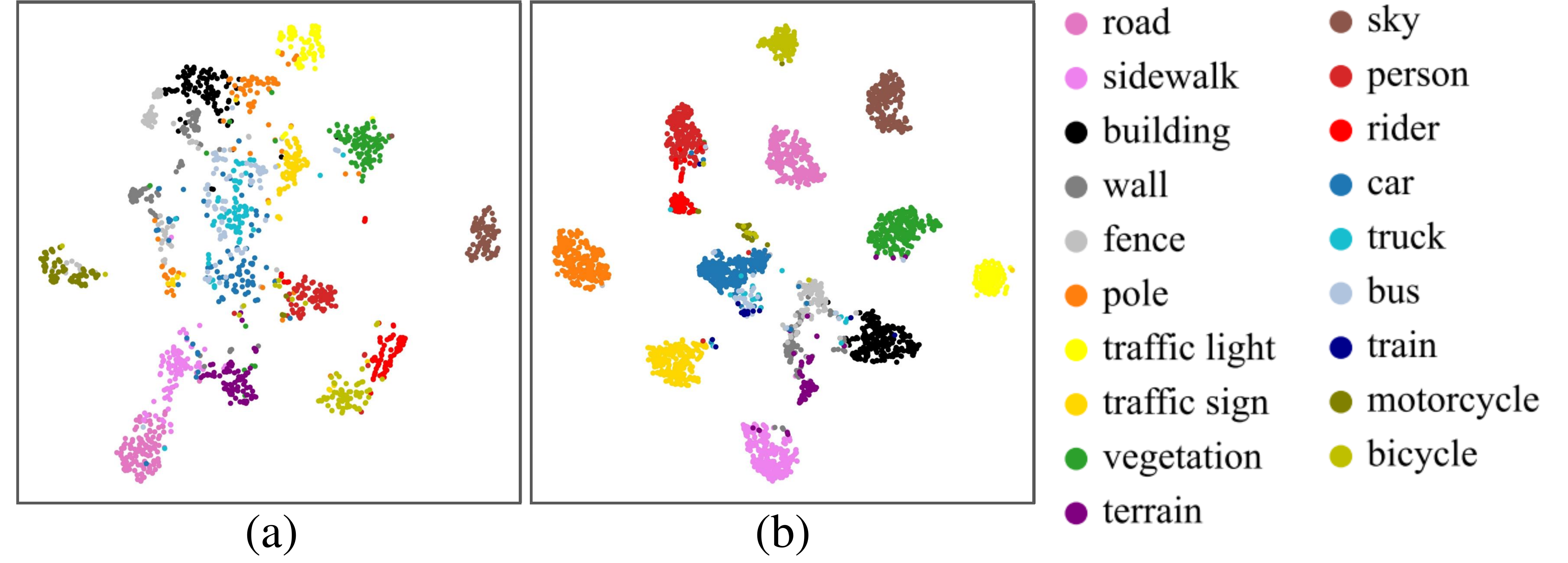}
\caption{t-SNE visualization for \textbf{(a) pixel-wise NPM method} and \textbf{(b) our proposed RAML method}. All learned metrics of 19 classes of the \textit{Cityscapes} dataset are included, where \textit{car}, \textit{truck} and \textit{bus} are OOD classes.}
\label{fig:tsne}
\end{figure}

\begin{table*}[htb]
\centering
\large
\resizebox{1.0\textwidth}{!}
{
\begin{tabular}{lcccccccccccccccccccccccc}
\hline
16+1 setting & Method & \rotatebox{90}{road} & \rotatebox{90}{sidewalk}  &  \rotatebox{90}{building} & \rotatebox{90}{wall} &\rotatebox{90}{fence} & \rotatebox{90}{pole} & \rotatebox{90}{traffic light} & \rotatebox{90}{traffic sign} & \rotatebox{90}{vegetation} & \rotatebox{90}{terrain} & \rotatebox{90}{sky} & \rotatebox{90}{person} & \rotatebox{90}{rider} & \rotatebox{90}{train} & \rotatebox{90}{motorcycle} & \rotatebox{90}{bicycle} &  \rotatebox{90}{car} &  \rotatebox{90}{truck} &  \rotatebox{90}{bus} & \rotatebox{90}{\textbf{mIoU\scriptsize{all}}} & \rotatebox{90}{\textbf{mIoU\scriptsize{novel}} }& \rotatebox{90}{\textbf{mIoU\scriptsize{old}} } & \rotatebox{90}{\textbf{mIoU\scriptsize{harm}} }  \\
\hline
\multirow{3}{*}{Baseline} & All 17 & 97.8 & 82.4 & 91.8 & 52.3 & 57.5 & 59.9 & 64.1 & 74.2 & 91.9 & 61.4 & 94.6 & 79.4 & 58.8 & 75.6 & 61.7 & 74.9 & 94.8 & - & - & 74.9 & - & - & - \\
 & First 16 & 98.0 & 82.1 & 91.4 & 43.6 & 56.4 & 58.9 & 61.4 & 72.6 & 91.6 & 60.5 & 94.4 & 79.1 & 57.6 & 67.9 & 61.1 & 75.1 & - & - & - & 72.0 & - & - & - \\
 & FT & 0.0 & 0.0 & 0.0 & 0.0 & 0.0 & 0.0 & 0.0 & 0.0 & 0.0 & 0.0 & 0.0 & 0.0 & 0.0 & 0.0 & 0.0 & 0.0 &  \cellcolor{SkyBlue}6.6 & - & - & 0.4 & 6.6 & 0.0 & 0.0\\
\hline
\multirow{3}{*}{5 shot} & \textbf{PLM} & 97.1 & 79.3 & 89.2 & 41.9 & 55.3 & 57.5 & 60.8 & 71.0 & 91.1 & 59.4 & 93.9 & 73.3 & 49.2 & 34.2 & 14.3 & 51.8 & \cellcolor{SkyBlue}75.7 & - & - & 64.4 & 75.7 & 63.7 & 69.2 \\
& \textbf{NPM} & 96.2 & 79.3 & 89.2 & 41.6 & 52.0 & 56.3 & 61.1 & 69.4 & 90.4 & 58.8 & 94.1 & 74.4 & 55.3 & 53.4 & 39.2 & 70.3 & \cellcolor{SkyBlue}64.6 & - & - & 67.4 & 64.6 & 67.6 & 66.1 \\
& \textbf{RAML(Ours)} & 97.3 & 82.6 & 91.4 & 51.0 & 57.2 & 59.2 & 65.5 & 74.4 & 91.7 & 63.9 & 94.7 & 79.1 & 59.1 & 23.7 & 52.1 & 72.3 & \cellcolor{SkyBlue}85.2 & - & - & \textbf{70.6} & \textbf{85.2} & \textbf{69.7} & \textbf{76.7} \\
\hline
\multirow{3}{*}{1 shot} & \textbf{PLM} & 96.8 & 77.1 & 89.6 & 41.4 & 48.7 & 53.2 & 60.3 & 64.5 & 90.3 & 55.6 & 94.3 & 59.1 & 43.6 & 39.5 & 12.0 & 35.7 & \cellcolor{SkyBlue}64.5 & - & - & 60.4 & 64.5 & 60.1 & 62.2 \\
& \textbf{NPM} & 95.9 & 79.2 & 88.8 & 41.3 & 50.5 & 56.0 & 61.0 & 69.1 & 90.2 & 58.6 & 94.1 & 73.6 & 55.1 & 49.7 & 37.4 & 69.6 & \cellcolor{SkyBlue}60.1 & - & - & 66.5 & 60.1 & 66.9 & 63.3 \\
& \textbf{RAML(Ours)} & 97.4 & 82.6 & 91.5 & 51.0 & 57.3 & 59.3 & 65.5 & 74.4 & 91.8 & 64.0 & 94.7 & 79.2 & 59.1 & 11.5 & 52.2 & 72.4 & \cellcolor{SkyBlue}85.5 & - & - & \textbf{70.0} & \textbf{85.5} & \textbf{69.0} & \textbf{76.4} \\
\hline
\hline
16+3 setting & & & & & & & & & & & & & & & & & & & & & & & & \\
\hline
\multirow{3}{*}{Baseline} & All 19 & 97.9 & 83.0 & 91.7 & 51.5 & 58.3 & 59.8 & 64.2 & 74.2 & 92.0 & 61.2 & 94.6 & 79.7 & 59.1 & 63.9 & 61.5 & 75.0 & 94.2 & 78.5 & 81.4 & 74.8 & - & - & - \\
 & First 16 & 98.0 & 82.1 & 91.4 & 43.6 & 56.4 & 58.9 & 61.4 & 72.6 & 91.6 & 60.5 & 94.4 & 79.1 & 57.6 & 67.9 & 61.1 & 75.1 & - & - & - & 72.0 & - & - & -\\
 & FT & 0.0 & 0.0 & 0.0 & 0.0 & 0.0 & 0.0 & 0.0 & 0.0 & 0.0 & 0.0 & 0.0 & 0.0 & 0.0 & 0.0 & 0.0 & 0.0 & \cellcolor{SkyBlue}0.0 &\cellcolor{SkyBlue} 0.0 &\cellcolor{SkyBlue} 0.4 & 0.0 & 0.1 & 0.0 & 0.0 \\
\hline
\multirow{3}{*}{5 shot}& \textbf{PLM} & 97.1 & 79.2 & 84.8 & 38.1 & 46.4 & 56.8 & 58.8 & 61.0 & 91.0 & 59.3 & 92.9 & 63.6 & 47.5 & 3.4 & 13.8 & 47.5 & \cellcolor{SkyBlue}67.0 & \cellcolor{SkyBlue}5.7 & \cellcolor{SkyBlue}12.0 & 54.0 & 28.2 & 58.8 & 38.1 \\
& \textbf{NPM} & 96.1 & 79.3 & 58.7 & 41.5 & 51.5 & 56.3 & 60.7 & 69.0 & 90.4 & 58.8 & 94.1 & 74.3 & 55.1 & 32.0 & 39.1 & 70.2 & \cellcolor{SkyBlue}55.7 & \cellcolor{SkyBlue}1.6 & \cellcolor{SkyBlue}21.0 & 58.2 & 26.1 & 64.2 & 37.1 \\
& \textbf{RAML(Ours)} & 97.3 & 82.6 & 91.1 & 50.6 & 57.2 & 59.1 & 65.5 & 74.1 & 91.7 & 64.0 & 94.7 & 79.0 & 58.9 & 3.7 & 52.2 & 72.3 & \cellcolor{SkyBlue}79.3 & \cellcolor{SkyBlue}9.7 & \cellcolor{SkyBlue}26.0 & \textbf{63.6} & \textbf{38.4} & \textbf{68.4} & \textbf{49.1} \\
\hline
\multirow{3}{*}{1 shot}& \textbf{PLM} & 96.8 & 75.2 & 49.0 & 33.1 & 31.4 & 48.0 & 33.2 & 44.6 & 89.7 & 55.3 & 23.0 & 42.1 & 32.8 & 5.3 & 8.0 & 27.7 & \cellcolor{SkyBlue}30.4 & \cellcolor{SkyBlue}0.7 & \cellcolor{SkyBlue}9.5 & 38.7 &13.5 & 43.4 & 20.6 \\
& \textbf{NPM} & 95.8 & 79.2 & 44.6 & 41.2 & 50.2 & 56.0 & 60.5 & 67.5 & 90.1 & 58.6 & 94.0 & 73.5 & 54.9 & 24.9 & 37.2 & 69.6 & \cellcolor{SkyBlue}54.5 & \cellcolor{SkyBlue}1.1 & \cellcolor{SkyBlue}22.0 & 56.6 & 25.9 & 62.3 & 36.5\\
& \textbf{RAML(Ours)} & 97.4 & 82.6 & 91.3 & 50.3 & 56.0 & 59.2 & 65.5 & 74.1 & 91.7 & 63.9 & 94.7 & 79.1 & 58.9 & 3.9 & 52.2 & 72.4 & \cellcolor{SkyBlue}80.9 & \cellcolor{SkyBlue}5.5 & \cellcolor{SkyBlue}23.0 & \textbf{63.2} & \textbf{36.5} & \textbf{68.3} & \textbf{47.5} \\
\hline

\end{tabular}
}
\caption{Incremental few-shot learning results on \textit{Cityscapes} for 16+1 setting (OOD class is \textit{car}) and 16+3 setting (OOD classes are \textit{car}, \textit{truck}, \textit{bus}). The unknown classes are in \colorbox{SkyBlue}{blue}. Finetune (FT) is the baseline with catastrophic forgetting. }
\label{tab:open world segmentation}
\end{table*}

\begin{figure}[h]
\centering
\includegraphics[width=0.5\textwidth]{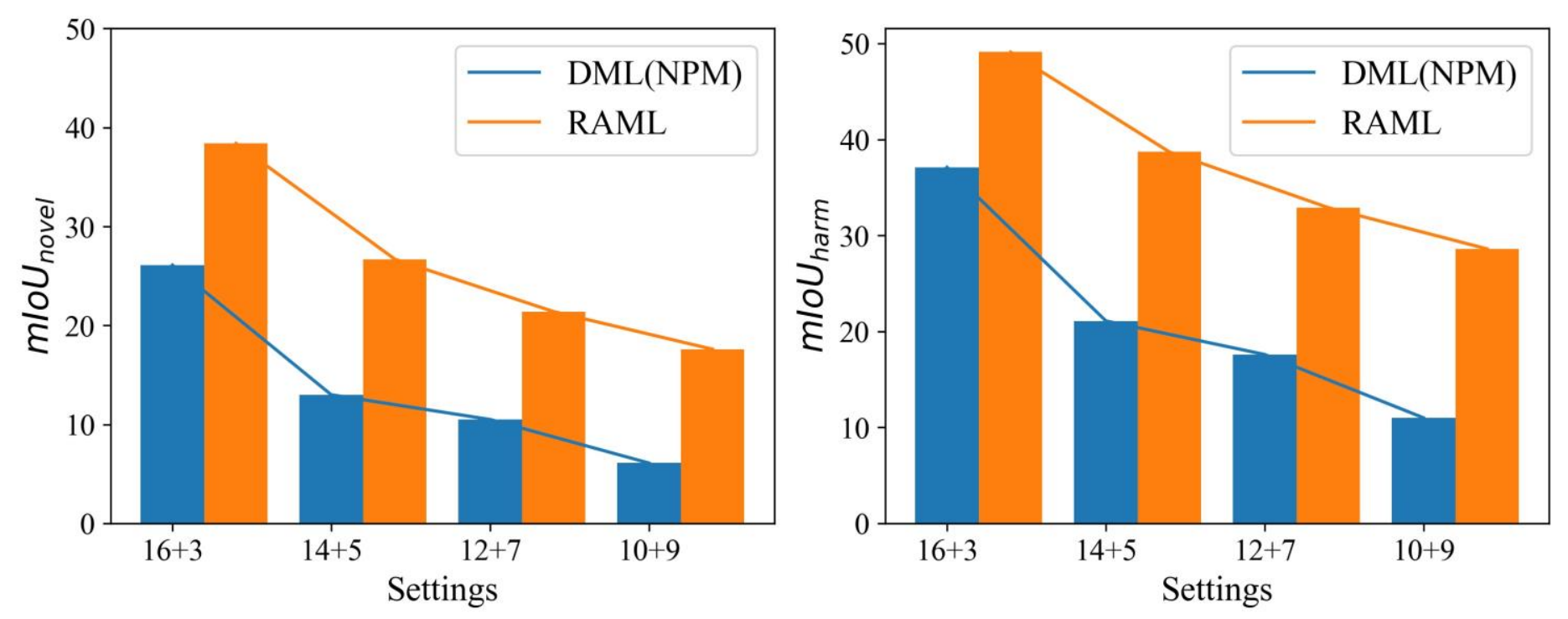}
\caption{Ablation study of the ratio of unknown classes to known classes. We compare our method to NPM and report results with \textbf{mIoU\scriptsize{novel}} and \textbf{mIoU\scriptsize{harm}}.}
\label{fig:seen}
\end{figure}

\begin{figure*}[htb]
\centering
\includegraphics[width=0.96\textwidth]{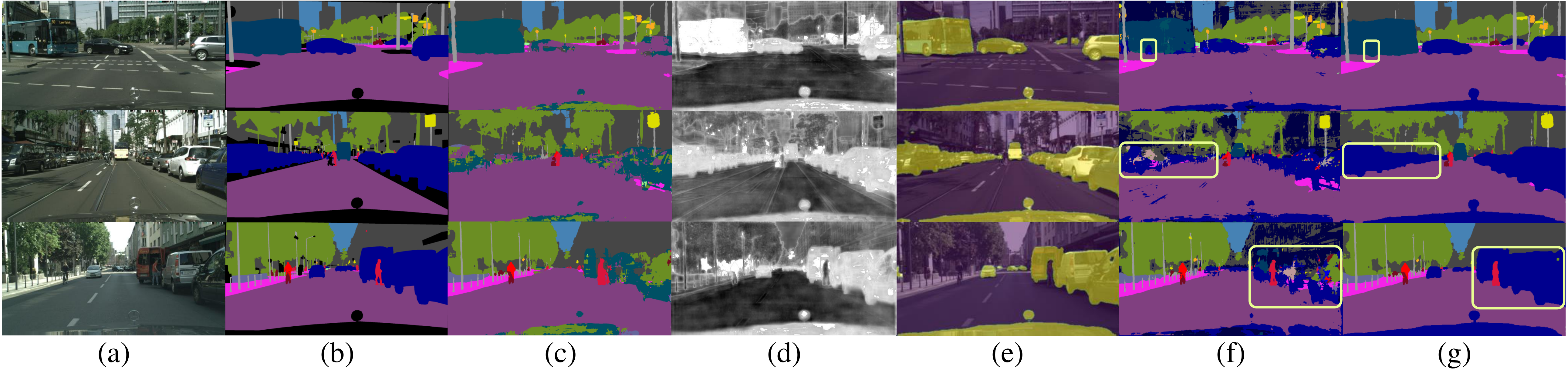}
\caption{Visual examples of RAML for \textbf{open world semantic segmentation}: (a) input images. (b) ground truth. (c) close-set outputs. 
(d) anomaly segmentation outputs. (e) MCA outputs. (f) results of pixel-wise NPM \protect\cite{cen2021deep}. (g) results of our RAML module. Yellow boxes indicate that RAML method can better ensure the integrity of the OOD objects. For example, in the first row, the pixel-wise method mistakenly divides the wheels of the bus into cars, while RAML can correctly segment the entire bus. (Best view in color and zoom in.)}
\label{fig:vis_all}
\end{figure*}

\paragraph{Results.} We test our method on \textit{CityScapes} and compare our method to pixel-wise \textbf{NPM} and \textbf{PLM} proposed by \cite{cen2021deep}. In our experiment, \textit{car}, \textit{truck}, and \textit{bus} are 3 OOD classes not involved in the training stage while the other 16 classes are regarded as in-distribution classes. As shown in \autoref{tab:open world segmentation}, our proposed RAML module outperforms the previous methods with a relatively large margin. According to \autoref{fig:vis_all}, pixel-wise metric learning shows erroneous broken segmentation results on OOD objects, while the proposed RAML demonstrates a remarkable ability to maintain the integrity of these results. In addition, \autoref{fig:tsne} shows that the feature embeddings produced by the proposed RAML maintain a reasonable inter-class distance and their intra-class distributions are also more concentrated. Such feature distribution could foster the model to obtain a robust decision boundary. 

\subsection{Ablation Study}
\label{sec:ablation}
\paragraph{Ratio of unknown classes to known classes.} The performance of the trained segmentation model has highly correlated with the amount of training information. We compare our proposed RAML method with the current SOTA method, NPM \cite{cen2021deep}, under the different ratios of unknown classes to known classes. As shown in \autoref{fig:seen}, although our RAML method has a decline in performance as the ratio increases, it outperforms NPM in all ratio settings.

\begin{table}[h]
	\begin{center}
	\resizebox{0.48\textwidth}{!}
    {
			\begin{tabular}{l|cccc}
				\hline
				Method & mIoU\scriptsize{all} & mIoU\scriptsize{novel} & mIoU\scriptsize{old} & mIoU\scriptsize{harm}\\ \hline
				Baseline & 49.1 & 1.5 & 58.0 & 2.9\\
				$+L_{rec}$ & 61.8 & 33.6 & 67.1 & 43.2\\
				$+L_{rec}+L_{split}$ & 62.6 & 37.6 & 67.3 & 48.3\\
				$+L_{rec}+L_{split}+L_{inter}$ & 63.6  & 38.4 & 68.3 & 49.1\\
				\hline
		\end{tabular}
	}
	\caption{Ablation study of losses used in MCA Module. Baseline is using Close-set Module directly.}
	\label{table:meta_loss}
	\end{center}
\end{table}

\paragraph{Losses in MCA.} This section evaluates the losses of our MCA module. As shown in \autoref{table:meta_loss}, the reconstruction loss ensures that our model obtains all information for the unknown classes, significantly improving the validity of MCA. The intersection loss and split loss also bring relatively smaller gains by improving the distribution of candidate regions in meta channels.

\section{Conclusion}
We have proposed RAML to enhance the performance of open world semantic segmentation. The main reason is that the region-aware feature outperforms the pixel-wise feature on maintaining the semantic integrity of the segmented OOD objects. Effective region separation methods are needed to realize RAML on anomaly segmentation and incremental few-shot learning. We, therefore, adopt the classic uncertainty-based methods to extract candidate regions for anomaly segmentation and propose an MCA module to further separate the anomaly regions for incremental few-shot learning. Experimental results show that our proposed method achieves the SOTA performance on the anomaly segmentation and the overall open world semantic segmentation. Our method has the potential to boost the use of open world semantic segmentation in practical applications.

\newpage
\section{Acknowledgments}
This work is supported by the Grants under the National Natural Science Foundation of China (NSFC) under Grants 12090022, 11831002, 71704023, and Beijing Natural Science Foundation (Z180001).
\bibliographystyle{named}
\bibliography{ijcai22}

\begin{thebibliography}{}

\bibitem[\protect\citeauthoryear{Baur \bgroup \em et al.\egroup
  }{2018}]{baur2018deep}
Christoph Baur, Benedikt Wiestler, Shadi Albarqouni, and Nassir Navab.
\newblock Deep autoencoding models for unsupervised anomaly segmentation in
  brain mr images.
\newblock In {\em MICCAI Brainlesion Workshop}, pages 161--169, 2018.

\bibitem[\protect\citeauthoryear{Bendale and Boult}{2015}]{7298799}
Abhijit Bendale and Terrance Boult.
\newblock Towards open world recognition.
\newblock In {\em CVPR}, pages 1893--1902, 2015.

\bibitem[\protect\citeauthoryear{Cen \bgroup \em et al.\egroup
  }{2021}]{cen2021deep}
Jun Cen, Peng Yun, Junhao Cai, Michael~Yu Wang, and Ming Liu.
\newblock Deep metric learning for open world semantic segmentation.
\newblock In {\em ICCV}, pages 15333--15342, 2021.

\bibitem[\protect\citeauthoryear{Chan \bgroup \em et al.\egroup
  }{2021}]{chan2021entropy}
Robin Chan, Matthias Rottmann, and Hanno Gottschalk.
\newblock Entropy maximization and meta classification for out-of-distribution
  detection in semantic segmentation.
\newblock In {\em ICCV}, pages 5128--5137, 2021.

\bibitem[\protect\citeauthoryear{Chen \bgroup \em et al.\egroup
  }{2020}]{chen2020learning}
Guangyao Chen, Limeng Qiao, Yemin Shi, Peixi Peng, Jia Li, Tiejun Huang,
  Shiliang Pu, and Yonghong Tian.
\newblock Learning open set network with discriminative reciprocal points.
\newblock In {\em ECCV}, pages 507--522, 2020.

\bibitem[\protect\citeauthoryear{Gal and Ghahramani}{2016}]{gal2016dropout}
Yarin Gal and Zoubin Ghahramani.
\newblock Dropout as a bayesian approximation: Representing model uncertainty
  in deep learning.
\newblock In {\em ICML}, pages 1050--1059, 2016.

\bibitem[\protect\citeauthoryear{He \bgroup \em et al.\egroup
  }{2016}]{he2016deep}
Kaiming He, Xiangyu Zhang, Shaoqing Ren, and Jian Sun.
\newblock Deep residual learning for image recognition.
\newblock In {\em CVPR}, pages 770--778, 2016.

\bibitem[\protect\citeauthoryear{Hendrycks and
  Gimpel}{2017}]{hendrycks2016baseline}
Dan Hendrycks and Kevin Gimpel.
\newblock A baseline for detecting misclassified and out-of-distribution
  examples in neural networks.
\newblock In {\em ICLR}, 2017.

\bibitem[\protect\citeauthoryear{Hendrycks \bgroup \em et al.\egroup
  }{2019}]{hendrycks2019scaling}
Dan Hendrycks, Steven Basart, Mantas Mazeika, Mohammadreza Mostajabi, Jacob
  Steinhardt, and Dawn Song.
\newblock Scaling out-of-distribution detection for real-world settings.
\newblock {\em arXiv preprint arXiv:1911.11132}, 2019.

\bibitem[\protect\citeauthoryear{Hu \bgroup \em et al.\egroup
  }{2021}]{hu2021region}
Hanzhe Hu, Jinshi Cui, and Liwei Wang.
\newblock Region-aware contrastive learning for semantic segmentation.
\newblock In {\em ICCV}, pages 16291--16301, 2021.

\bibitem[\protect\citeauthoryear{Joseph \bgroup \em et al.\egroup
  }{2021}]{joseph2021open}
K~J Joseph, Salman Khan, Fahad~Shahbaz Khan, and Vineeth~N Balasubramanian.
\newblock Towards open world object detection.
\newblock In {\em CVPR}, 2021.

\bibitem[\protect\citeauthoryear{Lis \bgroup \em et al.\egroup
  }{2019}]{lis2019detecting}
Krzysztof Lis, Krishna Nakka, Pascal Fua, and Mathieu Salzmann.
\newblock Detecting the unexpected via image resynthesis.
\newblock In {\em ICCV}, pages 2152--2161, 2019.

\bibitem[\protect\citeauthoryear{Oreshkin \bgroup \em et al.\egroup
  }{2018}]{oreshkin2018tadam}
Boris~N Oreshkin, Pau Rodriguez, and Alexandre Lacoste.
\newblock Tadam: Task dependent adaptive metric for improved few-shot learning.
\newblock In {\em NeurIPS}, 2018.

\bibitem[\protect\citeauthoryear{Pinggera \bgroup \em et al.\egroup
  }{2016}]{pinggera2016lost}
Peter Pinggera, Sebastian Ramos, Stefan Gehrig, Uwe Franke, Carsten Rother, and
  Rudolf Mester.
\newblock Lost and found: detecting small road hazards for self-driving
  vehicles.
\newblock In {\em IROS}, pages 1099--1106, 2016.

\bibitem[\protect\citeauthoryear{Saito \bgroup \em et al.\egroup
  }{2021}]{saito2021learning}
Kuniaki Saito, Ping Hu, Trevor Darrell, and Kate Saenko.
\newblock Learning to detect every thing in an open world.
\newblock {\em arXiv preprint arXiv:2112.01698}, 2021.

\bibitem[\protect\citeauthoryear{Schroff \bgroup \em et al.\egroup
  }{2015}]{Schroff_2015_CVPR}
Florian Schroff, Dmitry Kalenichenko, and James Philbin.
\newblock Facenet: A unified embedding for face recognition and clustering.
\newblock In {\em CVPR}, 2015.

\bibitem[\protect\citeauthoryear{Sobel and Feldman}{1968}]{sobel19683x3}
Irwin Sobel and Gary Feldman.
\newblock A 3x3 isotropic gradient operator for image processing.
\newblock {\em a talk at the Stanford Artificial Project in}, pages 271--272,
  1968.

\bibitem[\protect\citeauthoryear{Sun \bgroup \em et al.\egroup
  }{2020}]{2020Circle}
Y.~Sun, C.~Cheng, Y.~Zhang, C.~Zhang, L.~Zheng, Z.~Wang, and Y.~Wei.
\newblock Circle loss: A unified perspective of pair similarity optimization.
\newblock In {\em CVPR}, 2020.

\bibitem[\protect\citeauthoryear{Wang \bgroup \em et al.\egroup
  }{2021}]{wang2021exploring}
Wenguan Wang, Tianfei Zhou, Fisher Yu, Jifeng Dai, Ender Konukoglu, and Luc
  Van~Gool.
\newblock Exploring cross-image pixel contrast for semantic segmentation.
\newblock In {\em ICCV}, 2021.

\bibitem[\protect\citeauthoryear{Wen \bgroup \em et al.\egroup
  }{2016}]{wen_center}
Yandong Wen, Kaipeng Zhang, Zhifeng Li, and Yu~Qiao.
\newblock A discriminative feature learning approach for deep face recognition.
\newblock In {\em ECCV}, pages 499--515, 2016.

\bibitem[\protect\citeauthoryear{Xia \bgroup \em et al.\egroup
  }{2020}]{xia2020synthesize}
Yingda Xia, Yi~Zhang, Fengze Liu, Wei Shen, and Alan~L Yuille.
\newblock Synthesize then compare: Detecting failures and anomalies for
  semantic segmentation.
\newblock In {\em ECCV}, pages 145--161, 2020.

\bibitem[\protect\citeauthoryear{Xian \bgroup \em et al.\egroup
  }{2019}]{2019Semantic}
Y.~Xian, S.~Choudhury, Y.~He, B.~Schiele, and Z.~Akata.
\newblock Semantic projection network for zero- and few-label semantic
  segmentation.
\newblock In {\em CVPR}, 2019.

\bibitem[\protect\citeauthoryear{Yu \bgroup \em et al.\egroup
  }{2020}]{yu2020bdd100k}
Fisher Yu, Haofeng Chen, Xin Wang, Wenqi Xian, Yingying Chen, Fangchen Liu,
  Vashisht Madhavan, and Trevor Darrell.
\newblock Bdd100k: A diverse driving dataset for heterogeneous multitask
  learning.
\newblock In {\em CVPR}, pages 2636--2645, 2020.

\bibitem[\protect\citeauthoryear{Yuan \bgroup \em et al.\egroup
  }{2020}]{yuan2021segmentation}
Yuhui Yuan, Xiaokang Chen, Xilin Chen, and Jingdong Wang.
\newblock Segmentation transformer: Object-contextual representations for
  semantic segmentation.
\newblock In {\em ECCV}, 2020.

\bibitem[\protect\citeauthoryear{Zhao \bgroup \em et al.\egroup
  }{2017}]{zhao2017pyramid}
Hengshuang Zhao, Jianping Shi, Xiaojuan Qi, Xiaogang Wang, and Jiaya Jia.
\newblock Pyramid scene parsing network.
\newblock In {\em CVPR}, pages 2881--2890, 2017.

\bibitem[\protect\citeauthoryear{Zhong \bgroup \em et al.\egroup
  }{2021}]{Zhong_2021_CVPR}
Zhun Zhong, Linchao Zhu, Zhiming Luo, Shaozi Li, Yi~Yang, and Nicu Sebe.
\newblock Openmix: Reviving known knowledge for discovering novel visual
  categories in an open world.
\newblock In {\em CVPR}, pages 9462--9470, 2021.

\end{thebibliography}

\newpage
\appendix

\begin{center} 
{\centering\section*{Appendix}}
\end{center}

\renewcommand\thesection{\Alph{section}}
\renewcommand\thetable{\Alph{table}}
\renewcommand\thefigure{\Alph{figure}}

\section{Abbreviations}
\noindent\textbf{RAML}: \textbf{R}egion-\textbf{A}ware \textbf{M}etric \textbf{L}earning.

\noindent\textbf{MCA}: \textbf{M}eta-\textbf{C}hannel \textbf{A}ggregation.

\noindent\textbf{URS}: \textbf{U}ncertainty-based \textbf{R}egion \textbf{S}eparation.

\noindent\textbf{DMLNet}: \textbf{D}eep \textbf{M}etric \textbf{L}earning \textbf{Net}work.

\noindent\textbf{PML}: \textbf{P}ixel-wise \textbf{M}etric \textbf{L}earning.

\noindent\textbf{NPM}: \textbf{N}ovel \textbf{P}rototype \textbf{M}ethod.

\noindent\textbf{PLM}: \textbf{P}seudo \textbf{L}abel \textbf{M}ethod.

\noindent\textbf{OOD}: \textbf{O}ut-\textbf{O}f-\textbf{D}istribution.


\noindent\textbf{MSP}: \textbf{M}aximum \textbf{S}oftmax \textbf{P}robability.

\noindent\textbf{MaxLogit}: Maximum Logit.

\noindent\textbf{All 17}: Using 17 classes for full-supervised learning.

\noindent\textbf{All 19}: Using 19 classes for full-supervised learning.

\noindent\textbf{First 16}: Only using 16 known classes for close-set segmentation with full-supervised learning.

\noindent\textbf{FT}: Using unknown novel images for fine-tuning close-set segmentation model.

\section{Anomaly Segmentation}

\subsection{Details for expression 5}
For a new image $\mathbf{X}$ containing OOD objects, the uncertainty intensity Q is calculated as follows:
1) The close-set segmentation model infers the image $\mathbf{X}$ and produces the non-normalized logit $\mathbf{U}=\bar{\mathcal{G}}\circ \mathcal{F}(\mathbf{X})$ (Section 3.1). The small $\mathbf{U}$ value of a pixel means that it is more likely to belong to an unknown region.
2) We use a URS module to divide $\mathbf{X}$ into multiple regions, and the region-aware embeddings are calculated for each of them (expression (3)). 3) RAML computes the similarity between these region-aware embeddings and those of the known classes from the training set to obtain the maximum similarity $\mathcal{P}$ (expression (4)). If the maximum similarity is small, the region does not belong to any known classes. 4) Combining the results of 1) and 3), we take the product of $\mathbf{U}$ and $\mathcal{P}$. The negative of this product, named uncertainty intensity $\mathbf{Q}$, is thus positively related to the probability that a certain region is unknown.

\subsection{Implementation Details}
We use PyTorch (version 1.8.2) to implement our model and run it in the environment of CUDA 11.0. For anomaly segmentation, we follow~\cite{hendrycks2019scaling,cen2021deep} to use PSPNet~\cite{zhao2017pyramid} with ResNet101~\cite{he2016deep} as the close-set segmentation module. We set the batch size to 6 and train the model on two RTX-3090s in parallel. Moreover, we follow~\cite{cen2021deep} to train the module using SGD as the optimizer with a momentum of 0.9, a learning rate of $2\times10^{-2}$, and a learning rate decay of $10^{-4}$. 

We build the RAML module by applying two fully connected layers (4846 units and 128 units, respectively) after the close-set segmentation module, as shown in Figure 2. We train the module for 1500 iterations with a batch size of 6. Other training parameters are the same as the close-set segmentation module mentioned above.

In addition, Sobel filtering \cite{sobel19683x3} compensates the low sensitivity of MSP \cite{hendrycks2016baseline} on the edges of small objects. In the experiments, we set the hyper-parameters $\alpha$ and $\beta$ in Equation 2 as 50 and 0.7 for \textit{Lost and Found}, and 150 and 0.4 for \textit{Road Anomaly}.

\section{Incremental Few-shot Learning}

This section introduces implementation details and more ablation studies of the incremental few-shot learning of our proposed RAML method.

\subsection{Implementation Details}

We follow \cite{cen2021deep} to use a DeeplabV3+ model as the backbone of our close-set model. In the MCA module, We set $K=4$, $\eta = 0.02$, $\lambda_{inter}=0.1$, $\lambda_{split}=0.1$ and $\lambda_{rec}=0.01$. In our metric leanring, we set $\theta_{novel}=0.8$ as the cosine threshold for the unknown objects, $N_1=256,N_2=128$ in embedding space, $m=0.25$ and $\gamma = 8$ in circle loss \cite{2020Circle}. 

We train the close-set segmentation model for $3\times{10}^{4}$ iterations. After the anomaly segmentation, we finetune the close-set segmentation model for another ${10}^{4}$ iterations with all training samples plus the few-shot novels. We then build the RAML module by applying two fully connected layers (256 units and 128 units, respectively) after the MCA module, as shown in Figure 2. The RAML module is trained for $10^4$ iterations. 

In all training stages, we use an SGD optimizer with a momentum of 0.9, a learning rate decay of $10^{-4}$, and an initial learning rate of 0.01 for the feature extractor and 0.1 for the label predictor, respectively. Poly learning rate scheduler is used with the power of 0.9 for one epoch. Due to the GPU memory limitation, we set crop size to 762 in the close-set segmentation model and MCA module, and we further resize the feature maps to 512 in the RAML module. The batch size is set to 6. 

\subsection{Implementation Details for MCA}
In this section, we introduce how MCA module works in the inference stage. First, we select \textbf{candiate meta channel set} $\mathbf{C}_{cand}$ from annotated set $(\mathbf{X},\mathbf{Y})$. $\mathbf{C}_{meta}^i \in \{0,1\}^{K \times H \times W}$ is the output mask for an annotated image $(\mathbf{x}_i,\mathbf{y}_i)$, where $\mathbf{x}_i$ is the input image and $\mathbf{y}_i \in \{0,1\}^{H \times W}$ is the output mask for the novel class. A meta channel $\mathbf{c}^i_j$ is a \textbf{candiate meta channel} of meta channel output $\mathbf{C}_{meta}^i$ when it satify:

\begin{equation}
\frac{\sum_{h,w}\mathbf{c}^i_{j,hw}}{\sum_{h,w}\mathbf{y}_{hw}} > \kappa.
\end{equation}
Here, we set $\kappa=0.1$. The candiate meta channels set $\mathbf{C}_{cand}$ for all $L$ annotated images is defined as:
\begin{equation}
\mathbf{C}_{cand} = \cup_{i=1}^{L} \mathbf{C}_{cand}^i.
\end{equation}
As discussed in Section3.3. MCA module tends to segment objects based on local semantic information. Therefore, the information for novel classes is contained in the the candiate meta channel set $\mathbf{C}_{cand}$. In the inference stage, we aggregate the sub-regions from $\mathbf{C}_{cand}$ and get the output mask $\mathbf{C_{out}}$ via the exclusive disjunction of all $n$ channels in $\mathbf{C}_{cand}$, which can be expressed as:
\begin{equation}
\mathbf{C_{out}} = \mathbf{1}-(\mathbf{1}-\mathbf{c}_1) \odot (\mathbf{1}-\mathbf{c}_2) \odot \cdots \odot (\mathbf{1}-\mathbf{c}_{n}). 
\end{equation}
Then we use a post-processing algorithms on $\mathbf{C}_{out}$, including the hole filling and connected component algorithms, to generate the candidate regions $\mathcal{R}$ described in Section3.3.
\subsection{An example for RAML module}
\begin{figure}
    \centering
    \includegraphics[width=0.48\textwidth]{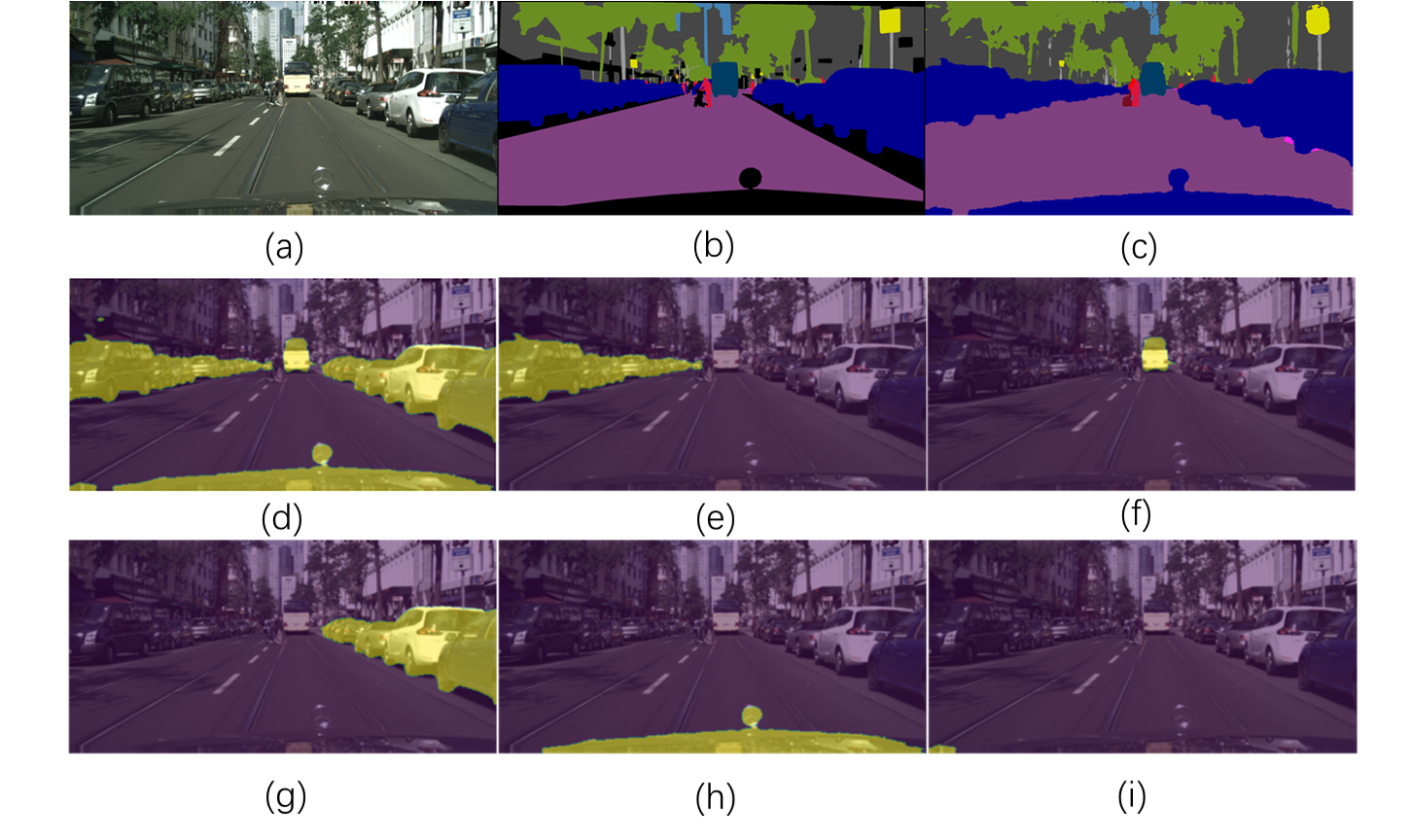}
    \caption{A Visual example of RAML Module in 16+3 setting. (a) Input image. (b) Ground truth. (c) Open World Semantic Segmentation output. (d) Aggregated Meta-channel. (e)-(i) Candidate regions.}
    \label{fig:RML_example}
\end{figure}

\autoref{fig:RML_example} shows an example for our proposed RAML method. In this case, MCA separates the anomaly regions into 5 candidate regions for the final stage of incremental few-shot learning. After that, the RAML module computes the region-aware feature embedding $f_{object}$ for each region according to Equation 3 and classifies it as a certain novel class according to Equation 12. As shown in \autoref{table:RML_example}, the RAML module correctly classifies the unknown classes even there’s more than one cosine similarity above the threshold $\theta_{novel}$. Besides, we notice that RAML can classify Regions (h) and (i) in \autoref{fig:RML_example} correctly which are ignored in the labels of \textit{CityScapes} dataset.

\begin{table}[h]
\begin{center}
\resizebox{0.4\textwidth}{!}
{
\begin{tabular}{c|cccc}
\hline
Candidate Region & Output & Car & Truck & Bus\\ 
\hline
(e) & Car & \textbf{0.97} & 0.82 & 0.92 \\
(f) & Bus & 0.87 & 0.75 & \textbf{0.93} \\
(g) & Car & \textbf{0.97} & 0.77 & 0.92 \\
(h) & Car & \textbf{0.91} & 0.65 & 0.69 \\
(i) & Car & \textbf{0.80} & 0.48 & 0.57 \\
\hline

\end{tabular}
}
\caption{Classification output for RAML module on the example in \autoref{fig:RML_example}. Cosine similarity is reported between each pair of candidate regions of the unknown classes in 16+3(car,truck,bus) settings where $\theta_{novel}=0.8$.}
\label{table:RML_example}
\end{center}
\end{table}

\subsection{Details for Circle Loss}
Circle loss was proposed by \cite{2020Circle}, which offers a unified formula for both two deep metric learning paradigms, i.e.,  learning with class-level labels and pair-wise labels. Meanwhile, it provides a  a more flexible optimization way towards a more definite convergence aim. Because of its superiority in metric learning, we adopt it in our RAML module.

Formally, given a region-aware feature embedding $f_{object}$, suppose that there are $K$ intra-class similarity scores, and $L$ inter-class similarity scores corresponding to $f_{object}$. We denote them as $\{s_{p}^{i}\}(i=1,2,\cdots,K)$ and $\{s_{n}^{i}\}(i=1,2,\cdots,L)$. The circle loss can be expressed as:
\begin{equation}
    \begin{aligned}
\mathcal{L}_{circle} &=\log \left[1+\sum_{i=1}^{K} \sum_{j=1}^{L} \exp \left(\gamma\left(s_{n}^{j}-s_{p}^{i}+m\right)\right)\right] \\
&=\log \left[1+\sum_{j=1}^{L} \exp \left(\gamma\left(s_{n}^{j}+m\right)\right) \sum_{i=1}^{K} \exp \left(\gamma\left(-s_{p}^{i}\right)\right)\right],
\end{aligned}
\end{equation}
where $\gamma$ is scale factor and $m$ is a margin for better similarity separation. We set $\gamma=8.0$ and $m=0.25$ in this work.

\subsection{Ablation study of region separation methods}
$K$ is the number of meta channels as discussed in Section 3.3. We compare our method with different $K$s to the pixel-wise methods and our proposed Uncertainty-based method (URS). As shown in \autoref{table:region separation}, region-aware methods perform consistently better than pixel-wise methods. The model performs best when $K=4$ and a reasonable explanation is that an overly small $K$ shrinks the benefit from the multiple meta channels over-segmenting the anomaly regions, but an overly large $K$ may cause excessive candidates for region separation, which are more challenging to be aggregated correctly.

\begin{table}[h]
\begin{center}
\resizebox{0.48\textwidth}{!}
{
\begin{tabular}{l|cccc}
\hline
\textbf{16+3 settings}   & mIoU\scriptsize{all} & mIoU\scriptsize{novel} & mIoU\scriptsize{old} & mIoU\scriptsize{harm}\\ 
\hline
PLM\scriptsize{latest} & 19.3 & 17.1 & 19.7 & 18.3 \\
PLM\scriptsize{all} & 38.7 & 13.5 & 43.4 & 20.6 \\
NPM & 56.6 & 25.9 & 62.3 & 36.5 \\
\hline
URS & 60.4 & 26.7 & 66.7 & 38.1 \\
MCA ($K=2$) & 61.0 & 34.7 & 65.9 & 45.5 \\
MCA ($K=4$) & 63.2 & 36.5 & 68.3 & 47.5 \\
MCA ($K=6$) & 59.7 & 33.1 & 64.7 & 43.8 \\
\hline
\end{tabular}
}
\caption{Ablation study of region separation methods, including pixel-wise method, RAML with our proposed URS module, and RAML with our proposed MCA module with different numbers of meta channels.}
\label{table:region separation}
\end{center}
\end{table}


\begin{table}[h]
\begin{center}
\resizebox{0.48\textwidth}{!}
{
\begin{tabular}{lccccc}
\hline
\textbf{16+3 settings} & ${\theta}_{novel}$   & mIoU\scriptsize{all} & mIoU\scriptsize{novel} & mIoU\scriptsize{old} & mIoU\scriptsize{harm}\\ 
\hline
\multirow{5}{*}{5 shot} & 0.7 & 63.3 & 37.6 & 68.2 & 48.4\\
    & 0.75 & 63.5 & 38.1 & 68.3 & 48.9\\
    & 0.8 & 63.6 & 38.4 & 68.4 & 49.1 \\
    & 0.85 & 63.9 & 38.1 & 68.7 & 49.0\\
    & 0.9 & 64.1 & 37.2 & 69.2 & 48.4\\
\hline
\multirow{5}{*}{1 shot} &  0.7 & 63.2 & 36.1 & 68.3 & 47.2\\
    & 0.75 & 63.3 & 36.4 & 68.3 & 47.5\\
    & 0.8 & 63.3 & 36.5 & 68.3 & 47.5\\
    & 0.85 & 64.1 & 36.4 & 69.4 & 47.7\\
    & 0.9 & 63.3 & 32.5 & 69.1 & 44.2\\
\hline
\end{tabular}
}
\caption{Ablation study of cosine similarity threshold $\theta_{novel}$}
\label{table:cosine}
\end{center}
\end{table}

\subsection{Ablation study of cosine similarity threshold}

$\theta_{novel}$ controls the cosine similarity threshold for unknown classes as discussed in Section 3.3. As shown in \autoref{table:cosine}, large  $\theta_{novel}$ will improve the precision rate for unknown classes while reducing the recall rate on them (shown as the performance drop on known classes). To balance the performance on known classes and unknown classes, we select  $\theta_{novel}=0.8$.

\subsection{Ablation study of Novel selection}

As shown in \autoref{table:novel select}, the selection of novels affects the segmentation performance, which is more obvious in 1-shot settings. As only a few data are labeled as novels, their center embedding may drift heavily from the overall data distribution. It is worth noting that for a fair comparison, our results reported in Table 2 adopt a fixed way of selecting novels, which follows \cite{cen2021deep} to choose the images with the largest area of the unknown classes as novels.

\begin{table}[h]
\begin{center}
\resizebox{0.48\textwidth}{!}
{
\begin{tabular}{lcccc}
\hline
\textbf{16+1 settings}  & mIoU\scriptsize{all} & mIoU\scriptsize{novel} & mIoU\scriptsize{old} & mIoU\scriptsize{harm}\\ 
\hline
5 shot & 70.53 $\pm$ 0.32 & 85.51 $\pm$ 0.32 & 69.59 $\pm$ 0.30 & 76.75 $\pm$ 0.30 \\
1 shot & 70.28 $\pm$ 0.51 & 84.43 $\pm$ 1.58 & 69.52 $\pm$ 0.49 & 76.32 $\pm$ 0.38 \\
\hline
\textbf{16+3 settings}  &  &  &  & \\ 
\hline
5 shot & 63.67 $\pm$ 0.17 & 38.4 $\pm$ 0.68 & 68.42 $\pm$ 0.10 & 49.19 $\pm$ 0.58 \\
1 shot & 63.13 $\pm$ 0.39 & 35.81 $\pm$ 2.23 & 68.28 $\pm$ 0.07 & 46.93 $\pm$ 1.92 \\
\hline
\end{tabular}
}
\caption{Ablation study of novel selection. We randomly select 1 or 5 novels for 5 times and report mean and std of the \textbf{mIoU}s.}
\label{table:novel select}
\end{center}
\end{table}

\end{document}